\def\year{2021}\relax
\documentclass[letterpaper]{article} 
\usepackage{aaai21}  
\usepackage{times}  
\usepackage{helvet} 
\usepackage{courier}  
\usepackage[hyphens]{url}  
\usepackage{graphicx} 
\usepackage{natbib}  
\usepackage{caption} 
\usepackage{amsmath}
\usepackage{amssymb}
\usepackage{bm}
\usepackage{multirow}
\usepackage{floatrow}
\newfloatcommand{capbtabbox}{table}[][\FBwidth]
\newcommand*\samethanks[1][\value{footnote}]{\footnotemark[#1]}

\title {Context-Aware Graph Convolution Network for Target Re-identification}
\author {
    Deyi Ji\textsuperscript{\rm1}\thanks{Equal contribution.},
    Haoran Wang\textsuperscript{\rm2}\samethanks,
    Hanzhe Hu\textsuperscript{\rm3},
    Weihao Gan\textsuperscript{\rm1},
    Wei Wu\textsuperscript{\rm1},
    Junjie Yan\textsuperscript{\rm1}\\
}
\affiliations {
    \textsuperscript{\rm1} SenseTime Research~~~ 
    \textsuperscript{\rm2} Xidian University~~~ 
    \textsuperscript{\rm3} Peking University~~~ \\
    \{jideyi, ganweihao, wuwei, yanjunjie\}@sensetime.com,  wanghaoran@stu.xidian.edu.cn, huhz@pku.edu.cn
}

\begin{document}

\maketitle

\begin{abstract}
Most existing re-identification methods focus on learning robust and discriminative features with deep convolution networks. However, many of them consider content similarity separately and fail to utilize the context information of the query and gallery sets, e.g. probe-gallery and gallery-gallery relations, thus hard samples may not be well solved due to the limited or even misleading information. In this paper, we present a novel Context-Aware Graph Convolution Network (CAGCN), where the probe-gallery relations are encoded into the graph nodes and the graph edge connections are well controlled by the gallery-gallery relations. In this way, hard samples can be addressed with the context information flows among other easy samples during the graph reasoning. Specifically, we adopt an effective hard gallery sampler to obtain high recall for positive samples while keeping a reasonable graph size, which can also weaken the imbalanced problem in training process with low computation complexity.
Experiments show that the proposed method achieves state-of-the-art performance on both person and vehicle re-identification datasets in a plug and play fashion with limited overhead.
\end{abstract}

\section{Introduction}
\noindent Re-identification (Re-ID) problems aim to determine corresponding targets across multiple cameras or different locations in the same camera from the gallery set with a given probe. Existing Re-ID tasks focus on both person and vehicle which play an important role in intelligent surveillance systems. 
Much process has been made in recent years in solving the specific problems of illumination variations, viewpoint changes and occlusions \cite{feature_relate3,Zhou_2019_ICCV,Hou_2019_CVPR,Wang_2020_CVPR}. However, the study on the context of the query and gallery sets receives less attention in research community.
To further illustrate the point, we conduct an analysis of existing Re-ID methods.


On the one hand, a number of person Re-ID methods attempt to learn robust and discriminative features using well-designed deep CNN \cite{first_personreid_1,feature_relate1,feature_relate2,feature_relate3,feature_relate4,feature_relate5} or fine-grained loss functions \cite{loss_relate1,first_personreid_2,hermans2017defense,loss_relate2}, and have witnessed great performance improvements in recent years. However, they treat each probe-gallery pair separately, without considering the relations between the query and gallery images. Thus the context information has not been fully explored in these systems, as a result, many advanced works \cite{Zhou_2019_ICCV,abdnet,cama,Chen_2020_CVPR} achieved the high top-1 accuracy of almost 95\% on the Market-1501 dataset, but the mAP is still less than 90\%.

On the other hand, several methods have tried to optimize the ranking results by exploiting the context information in person Re-ID. However, most of them operate in an unsupervised or semi-supervised manner as a post-processing operation. Concretely, Karaman et al. \cite{karaman2014leveraging} used a discriminative Conditional Random Field (CRF) to exploit the local context among the near neighbors in a semi-supervised way. Garcia et al. \cite{garcia2015person} proposed a post-ranking framework and analyzed context information in an unsupervised way. K-reciprocal reranking \cite{reranking} adjusted the gallery ranks using Jaccard distance calculated with k-reciprocal nearest neighbors in the whole probe and gallery sets.
Generally, these methods have two weaknesses. One is that they are not able to take full advantage of context information among two sets, as they perform as a post-process without training under the supervision. Second, they are of the sizeable computational quantity, since they need to take all query and gallery data for the matrix calculation. The time and memory consumption issues are severe in practice, especially on a large scale dataset. Besides, it is impractical to obtain all query images in a real-life scenario, such as a surveillance camera where the probe persons are dynamically changing in real-time environments.

For the first time, Shen et al. \cite{sggnn} proposed a graph model to utilize the context information in a supervised way, named SGGNN. Their purpose was to learn a discriminative feature that can be used for binary classification. With a siamese network, they processed one query image and one gallery image to obtain the probe-gallery feature, then took these features as nodes and used a Graph Neural Network to refine further.
The work provided an empirical approach to explore context information, however, there are three main limitations that have not yet been addressed.
First, they neglect the differentia in the gallery set, randomly sample the gallery images to form probe-gallery nodes. Thus hard positive and negative ones are not explicitly studied without appropriate sample selection.
Second, they build the graph in a fully-connected way, where one node connects to all the other ones. This may incur redundant or even harmful connections between nodes. Actually, the method is essentially a full aggregation of node features, the gallery-gallery relations are used as the aggregate weights rather than to construct a meaningful graph.
Third, the SGGNN only used two shallow fully-connected layers for message passing, thus the resulting features may be inadequate for discriminative representation. In conclusion, the probe-gallery and gallery-gallery relations are not well studied in the graph-based network.

Recently, a considerable amount of models has arisen for reasoning on graph-structured data in Re-ID tasks. Zhou et al. \cite{zhou2018graph} proposed a patch-wise graph matching to handle spatial misalignment caused by large variations in viewpoints and human poses. Wu et al. \cite{wu2019unsupervised} learnt the underlying view-invariant representations by the unsupervised graph association. Though the graph structure fits well to the relation reasoning, these methods only utilized the graph to explore the content similarity of the probe-gallery pairs, the context similarity still remained unsolved.
Meanwhile, Graph Convolution Network (GCN) is proposed for the problem of classifying nodes in the graph. The convolution operation defined over graphs is a weighted aggregation of target node features with its neighbor nodes. With the guidance of the adjacency matrix, we can explicitly specify the connections between certain samples. In this way, we intend to use GCN to learn the context information for relation reasoning, considering the Re-ID task as a node classification problem.  

Based on the above analysis, we propose a novel Context-Aware Graph Convolution Network (CAGCN) for Re-ID tasks. The framework takes one probe image and a certain number of candidate gallery images as inputs, and outputs the relation between the probe and each gallery image directly. In response to the above concerns, we discuss three main questions:

\noindent\textit{1. How to select sufficient and meaningful positive samples for the graph construction?}
In practical Re-ID tasks, the gallery set may be extensive, for one probe image, the positive and negative samples in the gallery may be heavily imbalanced distributed. We propose an intuitive and effective hard gallery sampler instead of randomly selecting candidate gallery images, which guarantees the quality and quantity of positive probe-gallery samples.

\noindent\textit{2. How to reason the relations between query and gallery sets validly?}
We build a context-aware graph to explicitly study two relations. On the one hand, we model the probe-gallery relations as node features in a high-dimension space. On the other hand, the gallery-gallery relations are modeled in the edges in a low-dimension space. Especially, we take the easy positive, hard positive and hard negative samples into consideration, making the graph meaningful and reasonable.
Moreover, we introduce a deep GCN model for better feature learning in the process of relation reasoning.

\noindent\textit{3. How to design a flexible and adjustable system?}
Sometimes we need to be more resource-efficient in terms of time, memory, or power. For example, a robot might need to make decisions within a time and power limit.
However, existing re-ranking methods have high computational cost, as they need to take all images for the similarity calculation, which is impractical when dealing with large-scale datasets. Fortunately, our method can be easily adjusted in sample number due to the proposed sampler and graph design. For example, we can only take one query image and a fixed number of gallery images according to the resource constraints.

Overall, our contribution is threefold:	
\begin{itemize}
\item We propose a novel Context-Aware Graph Convolution Network (CAGCN) for target re-identification tasks. To our knowledge, it is the first attempt to introduce the GCN for exploring context of the query and gallery sets in Re-ID tasks. Especially, we build a context-aware graph with relation guided edges, which explicitly model the relations in the two sets, guaranteeing the validity and effectiveness of information flow. Furthermore, the network is trained end-to-end in a supervised way.
\item The graph in our framework is constructed from an adjustable set of meaningful gallery candidates, which are selected by the proposed hard gallery sampler strategy. It ensures that we can select only a small number of samples to achieve good results, and the performance also shows a consistent improvement with the increase in resources.
\item The proposed framework is practical and can show in a plug-and-play fashion. Experiments show that the framework achieves state-of-the-art results on both person and vehicle benchmark datasets, i.e., Market-1501, DukeMTMC-ReID and VeRi-776. Notably, the mAP accuracy is significantly improved in all cases.
\end{itemize}

\section{Related Work}

\subsubsection{Re-identification.}
Re-identification (Re-ID) tasks focus on both person and vehicle, and have been extensively studied in the passed few years.
Mainstream approaches for person re-identification focus on three main aspects: learning a powerful and discriminative feature representation for person images \cite{first_personreid_1,feature_relate1,feature_relate4,feature_relate5,feature_relate3,wang2018parameter,p2net,cama,feature_relate9,9229188}, designing a fine-grained distance metric for measuring similarity between image pairs \cite{loss_relate1,loss_relate2,hermans2017defense,first_personreid_2}, or adopting some post-processing algorithms to refine the retrieval results in an offline way \cite{leng2015person,reranking,li2012common}. Feature learning based on deep convolution neural networks was first introduced in \cite{first_personreid_1,first_personreid_2}, nowadays the kind of methods focus on both global and part-patch information \cite{alignedreid,strong-baseline,loss_relate2,Hou_2019_CVPR,wei2019adversarial}.
For metric learning \cite{yu2017cross,relate_hardaware}, triplet based loss functions are widely used \cite{hermans2017defense}.
Vehicle Re-ID gains increasing attention recently and shares common frameworks with person Re-ID. Recent vehicle Re-ID methods \cite{vanet,aaver,pamtri,lou2019veri, Feng_2018_CVPR_Workshops} mainly focus on embedding spatial and temporal space  or viewpoints information. Our method is beneficial to both tasks.

\subsubsection{Graph convolution network.}

Recently GCN draws increasing attention and has proved to be very effective in many computer vision tasks, such as action recognition and node classification. Kipf et al. \cite{SEMI-SUPERVISED} proposed a simple and well-behaved layer-wise propagation rule for neural network models and demonstrated its effectiveness in semi-supervised classification tasks.
Veli{\v{c}}kovi{\'c} et al. \cite{gat} proposed a graph attention network which incorporates the self-attention mechanism into propagation step.
Wang et al. \cite{wang2019gcncluster} introduced GCN to face clustering tasks. Li et al. \cite{deepgcn} attempted to solve the vanishing gradient problem in deep GCNs. Hu et al. \cite{hu2020class} proposed a class-wise dynamic GCN for semantic segmentation.
However, GCN is few introduced to Re-ID tasks. Shen et al. \cite{sggnn} proposed a similarity guided graph neural network, the relations are transformed by two shallow fully-connected layers to update probe-gallery similarity with handcrafted weights, and there is no connection among galleries.
Chen et al. \cite{chen2020rrgccan} aimed to optimize the ranking results with graph models in the embedding set, and introduced the attention mechanism to enhance important channels.
By contrast, we are the first to propose to use GCN to explore the context of the query and gallery.
Moreover, Re-ID requires high recall for positive samples while reducing the amount of negative samples in graph nodes. Thus the sample strategy for graph nodes needs to be well designed, which has not been studied before.

\subsubsection{Context.}

Several works \cite{li2012common,karaman2014leveraging,garcia2015person,su2018cascaded,reranking,ijcai2020-77,su2020transferable} have utilized contextual information in their networks. Li et al. \cite{li2012common} developed a re-ranking model by analyzing the relative information and direct information of near neighbors of each pair of images. Karaman et al. \cite{karaman2014leveraging} combined discriminative models of person identity with a CRF to exploit the local manifold approximation induced by the nearest neighbor graph in feature space, while a graph topology defined by distances between all person images in feature space leveraged local support for label propagation in the CRF. In \cite{garcia2015person}, an unsupervised re-ranking model was learnt by jointly considering the content and context information in the ranking list.  Wang et al. \cite{reranking} utilized the whole datasets information and adopted a Jaccard distance to refine the gallery ranks.
Though these works successfully integrate information from k-nearest neighbors to refine the results, the misleading matches still exist without effective guidance of supervision information. To tackle this problem, we propose to explore the context directly in the training process, during the graph aggregation under the label supervision.

\begin{figure*}[!ht]
\begin{center}
\includegraphics[width=1.0\linewidth]{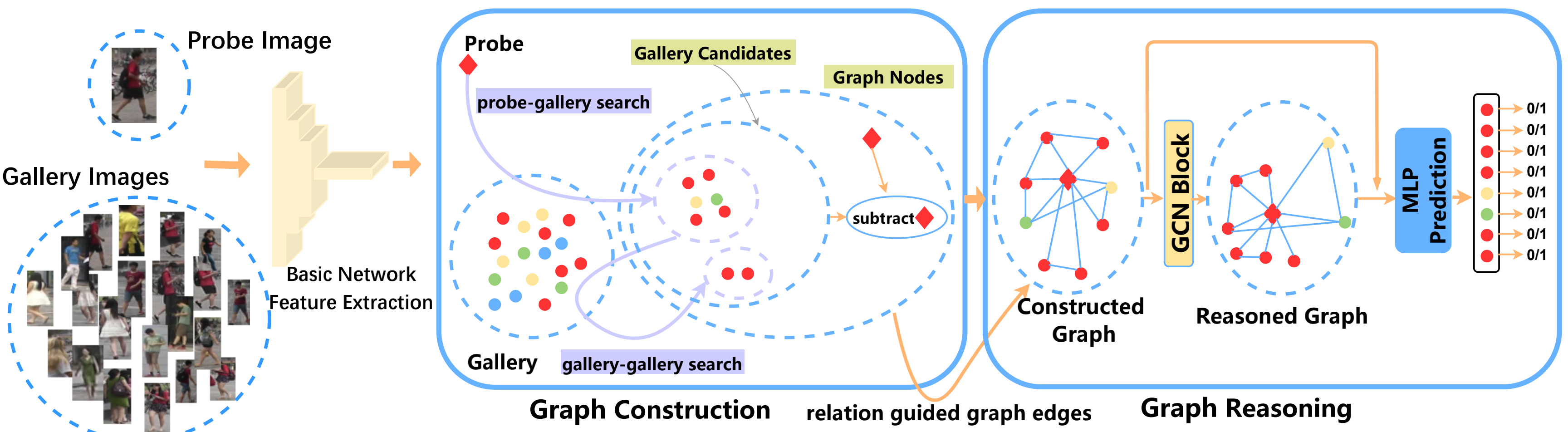}
\end{center}
\caption{An overview of our proposed framework. Given the probe and gallery images, we first feed them into the basic network for feature extraction. Then we utilize hard gallery sampler (HGS) method to select gallery nodes to construct the graph (different colors refer to different target identities), where probe-gallery search and gallery-gallery search take place to achieve high recall. Then we subtract the features of gallery candidates with probe feature to obtain node features. Then we construct the context-aware graph and GCN block for relation modeling. Finally, we perform an MLP on each node, denoting whether the corresponding gallery is identical to the probe.}
\label{framework}
\end{figure*}

\section{Approach}
In this section, we first give the problem definition, then we show the detail of the overall framework, context-aware graph and the process of reasoning on the graph.

\subsubsection{Problem definition.}

Given the probe image $p$ and gallery image set $G=\{g_i|i=1,2,3,...,n\}$, we use a basic network to extract their features, which are $\bm{f_p}$ and $\{\bm{f_{g_i}}|g_i\in G\}$. Then a hard gallery sampler is adopted to construct the gallery candidates set $G_c=\{g_i|i=1,2,3,...,k\}$. Subsequently, we obtain the probe-gallery features by subtracting gallery features from the probe feature $\bm{f_p}$, which are used as the node features for graph building.
The graph can be formulated as $G(V,E,A)$, where $V$, $E$ and $A$ denote the nodes, edges and adjacency matrix. $E$ and $A$ are carefully designed to build the context-aware graph. Finally, we apply graph convolutions on the built graph and then output the binary classification likelihood for each node, deciding whether the corresponding gallery is identical to the probe image.

\subsection{Framework Overview}
As illustrated in Figure~\ref{framework}, the proposed framework consists of three parts, including feature extraction, graph construction and graph reasoning. In particular, we adopt the model proposed in \cite{baseline} as the basic network. In the graph construction, probe-gallery search and gallery-gallery search select meaningful gallery candidates firstly, then the graph nodes are obtained by the subtraction operation of the probe-gallery pairs. The edges are explicitly designed with the guidance of gallery-gallery relations. In the graph reasoning, we propose the GCN block for a deeper network to fully explore the context information based on the built graph. Each node feature is aggregated with the context information from others under the label supervision. Finally, an MLP is adopted for binary classification for each graph node.
The framework can be applied to both person and vehicle Re-ID tasks in a plug and play fashion. 

\subsection{Context-Aware Graph Construction}
\label{graph_cons_sec}
As discussed above, for one specific probe, there may exist heavily imbalanced distribution of positive and negative samples in the gallery set. So during the graph construction, we first propose an effective hard gallery sampler to recall hard gallery samples and weaken the affect of imbalanced distribution. Then we show that in our GCN module, context information flow is well controlled with the relation guided graph edges, where hard probe-gallery features can be optimized with other easy probe-gallery features.

\begin{figure}[ht]
    \centering
    \includegraphics[width=1\linewidth]{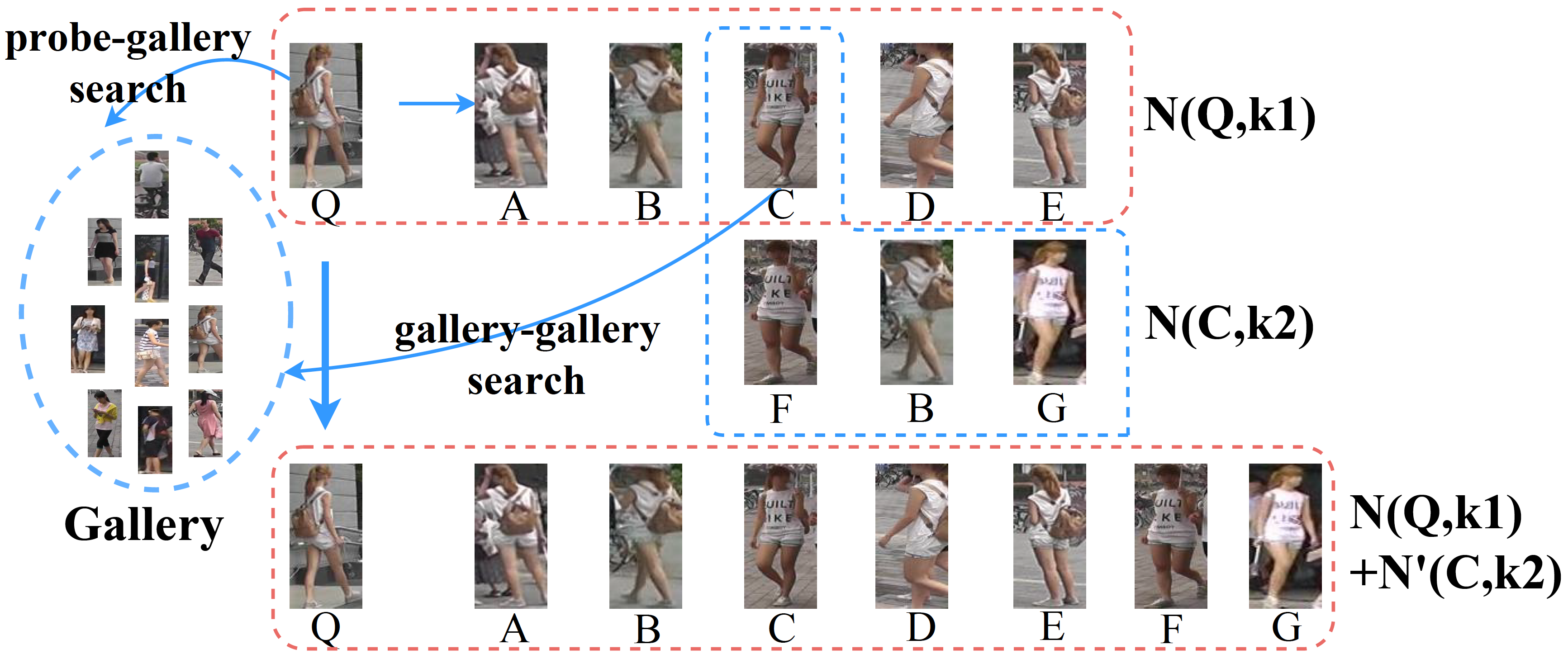}
    \caption{The proposed two-hop hard gallery sampler. For one specific probe  image $Q$, we first do probe-gallery search to select $k_1$-nearest  gallery images $N(Q, k_1)$. Then for each gallery image in $N(Q, k_1)$ (for example $C$), we perform gallery-gallery search to obtain $k_2$-nearest gallery images $N(C, k_2)$. Finally, we merge each gallery-gallery results to the probe-gallery results in order, until the total images number increases to $k$.}
    \label{HGS}
\end{figure}

\subsubsection{Hard gallery sampler.}
\label{sampler_sec}
In training stage, hard gallery samples are essential for context aggregation during graph reasoning, as they contain more discriminative information.
However, random sampling in the gallery set can not meet the requirements.
Thus, we design a Hard Gallery Sampler (HGS) to select hard positive gallery images. For each probe, we sample $k$ gallery candidates and denote them as $G_c=\{g_i|i=1, 2, 3...,k\}$.
A widely used plain sampler calculates the similarity between probe and gallery images and selects the $k$ nearest ones. However, in order to recall more hard positive samples, $k$ needs to be very large, which will introduce more negative samples and lead to more severe imbalanced distribution problems, as well as higher computation complexity.

To circumvent this issue, we propose an intuitive two-hop HGS method.
Though hard positive samples are usually not so similar as the probe in appearance due to the viewpoint changes, they can be retrieved by some easy positive ones as a bridge. Formally, given probe $p$ and gallery set $G$, we define $N(p,k)$ as $k$-nearest gallery images of $p$. In the sampling process, we first select the $k_1$-nearest gallery images $N(p,k_1)= \{g_1,g_2,g_3,...,g_{k1}\}$ and initialize $G_c = N(p, k_1)$. Then for each gallery image $g_i$ in $N(p, k_1)$, we select its $k_2$-nearest gallery images $N(g_i, k_2)$. Noted that there may be overlapped images between $N(p, k_1)$ and $N(g_i, k_2)$, we update $N(g_i, k_2)$ as:
\begin{equation}
    N'(g_i,k_2) = N(g_i,k_2)-N(g_i,k_2)\cap N(p,k_1).
\end{equation}
\noindent From $i=1$ to $i=k_1$, we add $N'(g_i,k_2)$ to $G_c$ until the total number of images in $G_c$ gets to $k$.
In general, both $k_1$ and $k_2$ are  smaller than $k$, thus both $N(p, k_1)$ and $N(g_i, k_2)$ have higher precision than $N(p, k)$. In this way, the hard positive samples can be obtained by the easy ones, so $G_c$ has a higher recall rate than the original $N(p,k)$.
The whole process is illustrated in Figure 2.

After obtaining $k$ gallery images by the HGS method, we get the node features $X$ of each probe-gallery pair by the subtraction operation, which explore and model the probe-gallery relations through the GCN.


\subsubsection{Relation guided graph edges.}

The most straightforward way is to construct a complete graph where any two nodes exist an edge. However, that may incur redundant or even harmful connection, passing messages between totally irrelevant nodes will finally lead to indistinguishable features. Actually, the method of SGGNN \cite{sggnn} can be viewed as a fully-connected graph, which may suffer the problem especially without sampling of gallery images.

The edge connection style in graph determines the context information aggregation flow among nodes. We analyze that in the retrieval task, a good information flow needs to ensure that, during graph reasoning, hard positive probe-gallery features (i.e. hard positive nodes) should be ``closer" to easy positive nodes, and vice versa. With the help of HGS, we solve this problem by desiging the meaningful edges between nodes. Specifically, given the gallery candidate set $G_c=\{g_i|i=1,2,3,...,k\}$, we first obtain $k'$ gallery neighbors of $g_i$ using the HGS method, denoted as $H(g_i)$. Then for another node $j$, if $g_j\in H(g_i)$, there will be an edge between node $i$ and node $j$. Formally, a sparsely connected graph is obtained by:
\begin{equation}
    \begin{aligned}
        &A_{ij}=\frac{\dagger (g_j\in H(g_i))exp(F(V_i,V_j))}{\sum_{j=1}^N{\dagger (g_j\in H(g_i))exp(F(V_i,V_j))}} \\
        &F(V_i,V_j)=\phi(V_i)^T\phi'(V_j) \\
    \end{aligned}
\end{equation}
where $A$ is the adjacency matrix, we perform softmax function on each node so that the sum of all the relation values of one node $i$ will be 1.
$\dagger(\cdot)$ is the indicator function, $\phi,\phi'$ denote two different linear transformations.
$F(V_i,V_j)$ denotes the gallery-gallery relations between two gallery samples with the dot-product similarity.

In this way, the context information aggregation flow is well controlled by the gallery-gallery relations. Without loss of generality, the sampled gallery candidate set $G_c$ is mainly composed of three types of nodes: easy positive, hard positive and hard negative nodes.
On one hand, during the edge connection process, hard positive nodes can be connected to easy positive ones, which helps narrow the gap between positive nodes, leading to the higher recall finally. On the other hand, for hard negative nodes, though sampled due to the vicinity to the probe image, the number of connections with positive nodes is limited by the graph formula, resulting in being pushed ``farther" from the positive nodes during graph reasoning under the label supervision.

Finally, we represent the constructed graph via a binary adjacency matrix $A \in \mathbb{R}^{|V|\times|V|}$, where $|V|$ is the node size.

\subsection{Graph Reasoning}
After the graph construction, context information can be aggregated among graph nodes with the supervision of each node label. The input of a graph convolution layer are the adjacency matrix $A$ and node feature matrix $X$, the reasoning process of one GCN layer can be denote as:
\begin{equation}
    \begin{aligned}
        Z^{(l+1)}=\sigma (AZ^lW^l)
    \end{aligned}
\end{equation}
where $A$ is the adjacent matrix. $Z^l \in R^{N \times d}$ is the feature representations of nodes in the $l$th layer, and $Z^0=X$. $W^l \in R^{d \times d'}$ is the layer-specific learnable weight matrix, $d$ and $d'$ are the input and output dimension. $\sigma(\cdot)$ denotes an activation function, and we adopt ReLU in this work.

Simply stacking multiple graph convolution layers would lead to the over-smoothing problems. In our framework, we introduce the residual connections \cite{deepgcn} to alleviate the problem. As a result, our network can be deeper with the GCN residual blocks.
One GCN residual block consists of one GCN layer followed by one dimension batchnorm layer. Formally, the output of one GCN residual block can be written as:
\begin{equation}
\label{resconnection}
\begin{aligned}
Z_{l+1} = Z_l + \mathcal{G}(Z_l, A).
\end{aligned}
\end{equation}

\noindent Where $\mathcal{G}$ denotes Eq. 3 followed by a dimension batchnorm layer.
Furthermore, since the graph convolution mainly focuses on the global context information during reasoning process, we concatenate the input feature to maintain both local and global information.
Each node feature represents a fully-explored probe-gallery pair, fused with context information in both query and gallery sets. Finally, we perform the binary classification on each node to decide whether the pair is the same person.

In the inference stage, the GCN module outputs the classification probability for each node, which can be viewed as their feature distance.
Moreover, the GCN module aims at learning probe-gallery and gallery-gallery relations, which focuses on the global information between the query and gallery sets.
While the original distance (calculated by origin visual feature) still requires attention since it also contains individual probe-gallery information that is conducive to the final feature representation.
For each $g_i$, we denote the original probe-gallery distance as $d_o(p, g_ i)$, and the predicted distance by the CAGCN as $d_g(p, g_i)$. Hence, the final distance is obtained by:
\begin{equation}
\label{dist_fusion}
\begin{aligned}
d(p, g_i) = d_o(p, g_i) + \lambda d_g(p, g_i),
\end{aligned}
\end{equation}

where  $\lambda$ controls the effect of the CAGCN. If $\lambda=0$, the final distance is just the original distance. By adding the distance from our method, we can achieve higher \textit{mAP} and preserve high \textit{Rank} metrics in all experiments.

\section{Experiments}
To verify the effectiveness of the proposed method, we conduct experiments on three large-scale Re-ID  datasets,  including two person datasets: Market1501 \cite{market} and DukeMTMC-reID \cite{duke}, and one vehicle dataset: VeRi-776 \cite{veri}.

\subsection{Implementation Details}

We utilize the network proposed in \cite{baseline} as baseline model and reproduce their results using the same settings, ResNet-50 is used as the backbone. The GCN module in our framework consists of 9 graph convolution layers followed by 3 fully connected layers. We use stochastic gradient descent to optimize our GCN module. The learning rate is 0.01 and the momentum is 0.9. The weight decay rate is 1e-4. The model is trained for 500 epochs with the batch size of 4. Focal loss is adopted as loss function, the $\alpha$ and $\gamma$ are set to 2 and 0.25. For the hard gallery sampler, we set $k_1$ to 70, $k_2$ to 20 and maximum nodes number $k$ to 100.
During the graph construction, only top-$8$ (i.e. $k'=8$) nodes are connected to construct a sparse graph.

\begin{table*}[!t]
\setlength{\abovecaptionskip}{0.cm}
\setlength{\belowcaptionskip}{-0.cm}
\centering
\small
\scalebox{1}{
\begin{tabular}{c|c|c|c|c|c|c|c}
\hline
\multirow{2}{*}{Methods} & \multirow{2}{*}{Reference}  & \multicolumn{3}{|c|}{Market1501} &  \multicolumn{3}{|c}{DukeMTMC-reID} \\
\cline{3-8}
~ &  & \textit{mAP} & \textit{Rank-1} & \textit{Rank-5} & \textit{mAP} & \textit{Rank-1} & \textit{Rank-5} \\
\hline
\hline
SFT \cite{sft} & ICCV 2019 & 82.7 & 93.4 & 97.4 & 73.2 & 86.9 & 93.9 \\
$P^2$Net \cite{p2net} & CVPR 2019 & 83.4	& 94 & 98 & 70.8 & 84.9 & 92.1 \\
BDB \cite{bdb} & ICCV 2019 & 84.3 & 94.2 & -  & 72.1  & 86.8 & -  \\		
CAMA \cite{cama} & CVPR 2019 & 84.5 & 94.7 & 98.1 & 72.9 & 85.8 & - \\
DCDS \cite{dcds} & ICCV 2019 & 85.8 & 94.8 & 98.1 & 75.5 & 87.5 & - \\
SCSN \cite{Chen_2020_CVPR} & CVPR 2020 & 88.5 & 95.7 & - & 79.0 & 90.1 & - \\
ISP \cite{Novotny_2020_ECCV} & ECCV 2020& 88.6 &95.3 & \textbf{98.6} & 80.0& 89.6 & \textbf{95.5} \\
\hline
CAGCN &  & \textbf{91.7} & \textbf{95.9} & 98.2 & \textbf{85.9} & \textbf{91.3} & 94.3 \\
CAGCN+RR & & 94 & 96.2 & 98 & 88.9 & 92.3 & 94.5 \\
\hline
\end{tabular}
}
\caption{Comparison results ($\%$) on Market1501 and DukeMTMC-reID dataset at 3 evaluation metrics: \textit{mAP}, \textit{Rank-1} and \textit{Rank-5}. The RR means the results with the re-ranking \cite{reranking} post-processing. The bold font denotes the best result without re-ranking.}
\label{market_sota}
\end{table*}
\subsection{Datasets and Evaluation Metrics}
The Market1501 dataset consists of 32,668 pedestrian image boxes of 1,501 identities, where 12,936 images from 751 identities are used as the training set, 3,368 probe images and 19,732 gallery images of remaining 750 identities are used as the testing set. The DukeMTMC-reID dataset contains 16,522 images of 702 identities for training, 2,228 probe images of the other 702 identities and 17,661 gallery images of 702 identities  for testing. The VeRi-776 dataset  consists of 49,357 images of 776 distinct vehicles captured by 20 non-overlapping cameras  from different viewpoints, resolutions and occlusions. We adopt the Cumulative Matching Characteristics (CMC) at \textit{Rank-1}, \textit{Rank-5}, \textit{Rank-10} and mean Average Precision (\textit{mAP}) to evaluate the performance of our model on all the datasets.

\subsection{Comparison with State-of-the-Art Methods}
We compare the proposed CAGCN with other methods on the above three datasets to demonstrate its effectiveness and generalization. Experiments show that our framework achieves the state-of-the-art performance on both person and vehicle tasks.
In Table \ref{market_sota}, we compare CAGCN with state-of-the-arts on Market1501 and DukeMTMC-reID. CAGCN achieves the best performance on both Rank-1 and \textit{mAP} evaluation metrics. Especially, our model surpasses the previous methods by a large margin on \textit{mAP}. The experimental results show that the proposed framework can exploit the context information among probe and gallery images effectively to learn more discriminative node features. When using the re-ranking scheme as post-processing, our method achieves the highest accuracy of 96.2\% and 94\% on Rank-1 and \textit{mAP}.
Table \ref{veri_sota} shows the results of CAGCN on VeRi-776 dataset compared with other state-of-the-art methods for vehicle re-identification. Our model also outperforms other methods on all the evaluation metrics, especially on \textit{mAP}.

%

\subsection{Ablation Study}
We utilize the network proposed in \cite{baseline} as baseline model and reproduce the results in our framework.

\subsubsection{The impact of different samplers.}
Table \ref{sampler} shows the performance of different sampling strategies on three datasets. Here we set the node and layer number in GCN module to 100 and 9 respectively. The plain sampler simply calculates the similarity between probe and gallery images and selects the $k$ nearest ones, while HGS refers to the proposed hard gallery sampler. Experiments show that a plain sampler can obtain most easy galleries and gain considerable improvements on all the metrics, especially \textit{mAP}. By applying hard gallery sampler, the \textit{mAP} can be further improved (1.2\% on Market1501, 2.3\% on DukeMTMC-reID and 1.1\% on VeRi-776), that is to say, more hard positive galleries are sampled and the recall rate of positive samples is further increased. It shows that the HGS effectively mitigates the affect of imbalanced problem discussed above. 



\begin{table}[!t]
\setlength{\abovecaptionskip}{0.cm}
\setlength{\belowcaptionskip}{-0.cm}
\small
\centering
\begin{tabular}{c|c|c|c}
\hline
Methods  & \textit{mAP} & \textit{Rank-1} & \textit{Rank-5}  \\
\hline
\hline
FDA-Net\cite{lou2019veri}  & 55.5 & 84.3 & 92.4 \\
VANet \cite{vanet} & 66.3 & 89.8 & 96.0  \\	
AAVER \cite{aaver}  & 61.2 & 89.0 & 94.7  \\	
PAMTRI \cite{pamtri}  &71.9 & 92.9 & 96.9  \\
PNVR \cite{pnvr}  & 74.3  & 94.3 & \textbf{98.7}  \\
\hline
CAGCN & \textbf{79.6} & \textbf{95.8} & \textbf{98.7} \\
\hline
\end{tabular}
\caption{Comparison results($\%$) on VeRi-776 dataset at 3 evaluation metrics: \textit{mAP}, \textit{Rank-1} and \textit{Rank-5}. Re-ranking is not used here. The bold font denotes the best result.}
\label{veri_sota}
\end{table}

\begin{table}[!ht]
\begin{center}
\begin{tabular}{c|c|c|c|c|c|c}
\hline
\multirow{2}{*}{Sampler} & \multicolumn{2}{|c|}{Market1501} &  \multicolumn{2}{|c|}{\begin{tabular}[c]{@{}l@{}} DukeMTMC \\  \ \ \ \ \ -reID \end{tabular}} &  \multicolumn{2}{|c}{VeRi-776} \\ \cline{2-7}

 &  \textit{mAP} & \textit{R-1}  & \textit{mAP} & \textit{R-1} & \textit{mAP} & \textit{R-1} \\
\hline
\hline
baseline & 85.4 & 94.1 & 75.6 & 86.3 & 74.1 & 93.9 \\
\hline
Plain & 90.4 & 95.0 & 83.1 & 89.1 & 78.1 & 94.7 \\
HGS & 91.6 & 95.0 & 85.4 & 89.6 & 79.2 & 95.2 \\
\hline
\end{tabular}
\end{center}
\caption{Ablation study of the effect of different samplers on the three datasets. \textit{R-1} is the omission of \textit{Rank-1}.}
\label{sampler}
\end{table}

%

\subsubsection{The impact of $\bm{\lambda}$ in final distance.}
\begin{figure}[ht]
    \centering
    \includegraphics[width=0.94\textwidth]{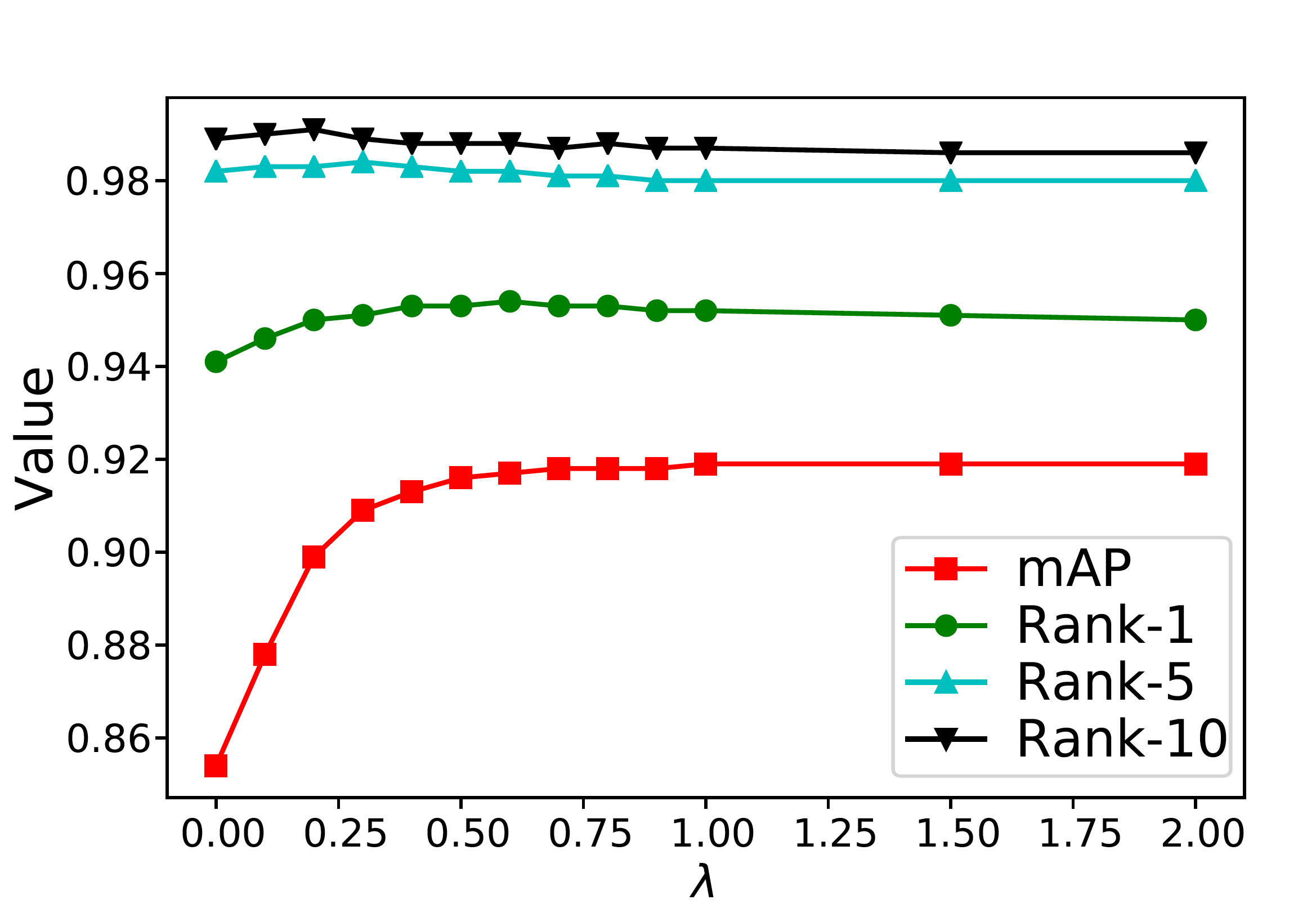}
    \caption{The impact of $\lambda$ in final distance on Market1501.}
    \label{lambda}
\end{figure}

$\lambda$ is used to balance the effect of two distances in Eq. 5. The greater the value, the greater the impact of CAGCN. As shown in Figure \ref{lambda}, with $\lambda$ gradually increasing, \textit{mAP} is gradually increasing and stay still around 0.92 when $\lambda$ gets to 1.0, which indicates that more and more context information is mined in this process. Meanwhile, \textit{Rank-1}, \textit{Rank-5} and \textit{Rank-10} are relatively stable, since the individual probe-gallery information is also preserved with the original distance, which has more influence 
on these indicators.



\subsubsection{The impact of graph node number.}
Table \ref{node_ab} shows the influence of nodes number in the GCN module on Market1501. Here the hard gallery sampler is adopted. Experiments show that the \textit{mAP} and \textit{Rank-1} grow with the increase of nodes number, since more positive galleries can be sampled with larger number of nodes. Noted that the \textit{mAP} does not gain much improvement when we increase node number from 100 to 500, that indicates the recall rate of positive samples is saturated, and only sampling 100 nodes can achieve the almost same performance as 500 ones. Actually, it demonstrates the outstanding effectiveness of the proposed sampler from another aspect. It also shows that we can dynamically adjust the size of the model, to balance the resource constraints and performance in practice.

\subsubsection{The impact of graph layer number.}
Table \ref{layer_ab} shows the influence of layer number in the GCN module on Market1501. The nodes number is 100. 
Benefiting from the residual connection, we can extend the network deeper to obtain more powerful node representations. We compare the results from shallow to deep network (layers=3, 6, 9). Experiments show the \textit{mAP} improves with the growth of layer numbers.

\subsubsection{The impact of \bm{$k'$} for edge connection.}
We show the impact of $k'$ during graph edge connection in Table \ref{graph_edge}, \textit{mAP} is basically stable when $k'$ is very small. However, if $k'$ is set too large, we find that \textit{mAP} decreases rapidly, since large $k'$ will lead to dense negative edge connections in the graph where the information aggregation is adversely affected.

\begin{table}[!ht]
\begin{center}
\begin{tabular}{c|c|c|c|c}
\hline
Node Number & \textit{mAP} & \textit{Rank-1} & \textit{Rank-5} & \textit{Rank-10} \\
\hline\hline
baseline & 85.4 & 94.1 & 98.2 & 98.9 \\
\hline
50 & 87.1 & 94.4 & 98.2 & 98.9 \\
80 & 88.3 & 94.4 & 98.3 & 98.9 \\
100 & 91.6 & 95.0 & 98.0 & 98.7 \\
500 & 91.7 & 95.9 & 98.2 & 98.8 \\
\hline
\end{tabular}
\end{center}
\caption{The performances of the GCN module with different nodes number on Market1501.}
\label{node_ab}
\end{table}

\begin{table}[!ht]
\begin{center}
\begin{tabular}{c|c|c|c|c}
\hline
Layer Number & \textit{mAP} & \textit{Rank-1} & \textit{Rank-5} & \textit{Rank-10} \\
\hline
\hline
baseline & 85.4 & 94.1 & 98.2 & 98.9 \\
\hline
3 & 88.4 & 94.1 & 98.0 & 98.8 \\
6 & 89.0 & 94.5 & 98.2 & 98.9 \\
9 & 91.6 & 95.0 & 98.0 & 98.7 \\
\hline
\end{tabular}
\end{center}
\caption{The performances of our GCN module with different number of layers on Market1501.}
\label{layer_ab}
\end{table}

\begin{table}[!ht]
\begin{center}
\begin{tabular}{c|c|c|c|c}
\hline
$k'$ & \textit{mAP} & \textit{Rank-1} & \textit{Rank-5} & \textit{Rank-10} \\
\hline
\hline
baseline & 85.4 & 94.1 & 98.2 & 98.9 \\
\hline
5 & 91.2 & 95.1 & 98.1 & 98.8 \\
8 & 91.6 & 95.0 & 98.0  & 98.7 \\
10 & 91.4 & 94.9 & 98.0 & 98.8 \\
15 & 91.0 & 94.9 & 98.1 & 98.9 \\
50 & 89.9 & 94.7 & 98.0 & 98.8 \\
\hline
\end{tabular}
\end{center}
\caption{The impact of $k'$ for graph edge connection on Market1501.}
\label{graph_edge}
\end{table}

%


\begin{figure}[ht]
    \centering
    \includegraphics[width=1\linewidth]{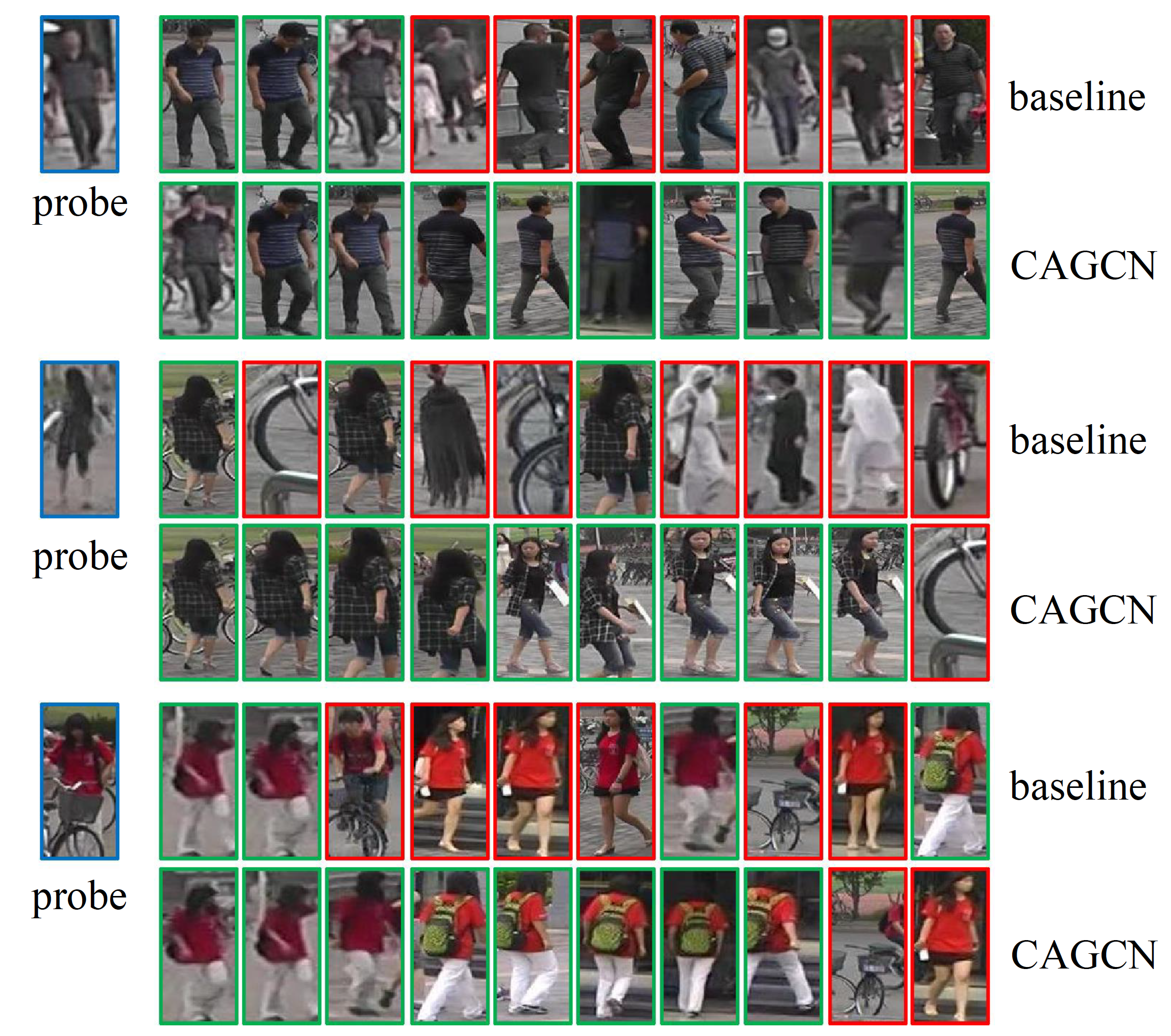}
    \caption{Qualitative results of three probes on Market1501. For each probe, the first row and second row correspond to the ranking results produced by baseline model and CAGCN framework. The blue, green and red box denotes the probe, true positive and false positive respectively.}
    \label{visual}
\end{figure}





\subsubsection{Qualitative results.}
Finally, we show the probe image and its related top-10 gallery images obtained by baseline model and CAGCN framework respectively. As shown in Figure \ref{visual}, CAGCN is able to address more hard positive samples and discard hard negative samples even if the probe images are in low quality. More comprehensively, we can notice that, despite that the person in the probe image is facing front, the framework can also retrieve the images where the person is facing back (line 2), and vice versa (line 4). Besides, the framework is more robust to color variance compared with the baseline (line 6). The same phenomenon can be found in other examples, and all cases illustrate that CAGCN can exploit the context information more effectively.


\section{Conclusion}

In this work, we propose a novel CAGCN framework based on deep graph convolution network to utilize the context information among probe and gallery images for target re-identification. CAGCN tackles the challenges existing in graph construction and effective information aggregation, and utilizes the novel hard gallery sampler to obtain high recall for positive samples and keep a reasonable graph size. Additionally, edges among graph nodes are controlled by the original gallery-gallery relations to avoid redundant connections. 
The proposed framework achieves state-of-the-art performances on both person and vehicle benchmarks.

\bibliography{egbib}
\end{document}